\documentclass[journal]{IEEEtran}

\usepackage{times}
\usepackage{epsfig}
\usepackage{graphicx}
\usepackage{amsmath}
\usepackage{amssymb}
\usepackage{enumitem}
\usepackage{array}
\usepackage{multirow}
\usepackage{graphicx}
\usepackage{subfigure}
\usepackage{amsthm,amsmath,amssymb}
\usepackage{mathrsfs}
\usepackage{booktabs}
\usepackage{cite}
\usepackage{color}

\ifCLASSINFOpdf
\else
\fi

\begin{document}

\title{Graph-Based Spatial-Temporal Convolutional Network for Vehicle Trajectory Prediction in Autonomous Driving}

\author{\IEEEauthorblockN{Zihao Sheng,
Yunwen Xu,~\IEEEmembership{Member, IEEE}, 
Shibei~Xue,~\IEEEmembership{Senior Member, IEEE}, and
Dewei Li}
\thanks{This work was supported in part by National Key Research and Development Project of China under Grant 2020YFB1600400, and in part by the National Natural Science Foundation of China under Grant 61973214, Grant 61873162 and Grant 62003210, and in part by the Natural Science Foundation of Shanghai under Grant 19ZR1476200, and in part by the Guangdong Key Research and Development Project under Grant 2020B0101050001. ({\it Corresponding author: Shibei Xue.})}
\thanks{Zihao Sheng, Yunwen Xu, Shibei Xue, and Dewei Li are with the Department of Automation, Shanghai Jiao Tong University and the Key Laboratory of System Control and Information Processing, Ministry of Education of China, Shanghai 200240, China (E-mails: \{zihaosheng, willing419, shbxue, dwli\}@sjtu.edu.cn).}}

\markboth{IEEE TRANSACTIONS ON INTELLIGENT TRANSPORTATION SYSTEMS}%
{Sheng \MakeLowercase{\textit{et al.}}: Graph-Based Spatial-Temporal Convolutional Network for Vehicle Trajectory Prediction in Autonomous Driving}

\maketitle

\begin{abstract}
Forecasting the trajectories of neighbor vehicles is a crucial step for decision making and motion planning of autonomous vehicles. 
This paper proposes a graph-based spatial-temporal convolutional network (GSTCN) to predict future trajectory distributions of all neighbor vehicles using past trajectories. 
This network tackles the spatial interactions using a graph convolutional network (GCN), and captures the temporal features with a convolutional neural network (CNN). The spatial-temporal features are encoded and decoded by a gated recurrent unit (GRU) network to generate future trajectory distributions.
Besides, we propose a weighted adjacency matrix to describe the intensities of mutual influence between vehicles, and the ablation study demonstrates the effectiveness of our proposed scheme.
Our network is evaluated on two real-world freeway trajectory datasets: I-80 and US-101 in the Next Generation Simulation (NGSIM).
Comparisons in three aspects, including prediction errors, model sizes, and inference speeds, show that our network can achieve state-of-the-art performance.
\end{abstract}


%
\IEEEpeerreviewmaketitle

\section{Introduction}
\IEEEPARstart{H}{uman} needs to continually observe nearby vehicles' behaviors during driving so as to plan future motions for safely and efficiently passing through complex traffics.
Similarly, autonomous vehicles should collect the movement information of nearby objects and then decide which maneuver can minimize risks and maximize efficiencies.
Detection of such information has become possible with the development of onboard sensors and vehicle-to-infrastructure (V2I) communication \cite{xie2017vehicle,dorrell2015connected}.
Recent works focus on how to utilize this information to plan motions for autonomous vehicles \cite{gonzalez2015review,claussmann2019review}.
One of the main aspects of motion planning is to predict other traffic participants' future trajectories \cite{mozaffari2020deep,rudenko2020human}, with which autonomous vehicles can infer future situations they might encounter \cite{li2015real,bae2020cooperation}.

In the past decades, inspired by vehicle evolution models \cite{lin2000vehicle} and statistics-based-models \cite{kang2017parametric,klingelschmitt2014combining}, research on vehicle trajectory prediction is undergoing developments, among which kinematic models \cite{brannstrom2010model,hillenbrand2006multilevel,polychronopoulos2007sensor} and Kalman filter \cite{mourllion2005kalman} have been widely studied.
These traditional models possess high computational efficiencies, such that they are especially suitable for real-time applications with limited hardware resources.
{However, if the spatial-temporal correlations of vehicles are ignored, the long-term prediction (e.g., longer than one second) calculated by these traditional methods would be unreliable \cite{lefevre2014survey}.}
{
In order to extend these traditional methods to consider the spatial-temporal interactions among vehicles, combination methods were proposed which take the advantages of traditional methods and advanced models (e.g., machine-learning-based models). 
For example, Ju {\it et al.} \cite{ju2020interaction} proposed a model which combines Kalman filter, kinematic models and neural network to capture the interactive effects among vehicles.
Since machine-learning-based models can model the interactions and learn nonlinear trajectory evolution from real-world data, these combination methods have shown better performance than the above approaches. }

So far, most of machine-learning-based methods have assumed that the task of vehicle trajectory prediction can be decomposed into two steps: firstly predict the maneuver of a target vehicle, and then generate a maneuver-based trajectory.
For example, Deo {\it et al.} \cite{deo2018would} used hidden Markov models to estimate maneuver intentions and then applied variational Gaussian mixture models for the trajectory predictions.
Tran {\it et al.} \cite{tran2014online} used a Gaussian process regression model to recognize the maneuver being performed by a target vehicle and applied Monte-Carlo method to predict future trajectories.
Other maneuver estimation models include support vector machines \cite{morris2011trajectory,aoude2012driver}, random forest \cite{schlechtriemen2015will}, multi-layer perceptrons \cite{ortiz2011behavior} and Bayesian networks \cite{lefevre2011exploiting,schreier2014bayesian}.
However, most of these approaches are highly dependent on handcrafted features, which are designed to model interactions and physical constraints among vehicles under certain scenarios.
Therefore, their performance would degrade in unexpected traffic scenarios.

As a branch of machine learning, deep learning can extract features automatically by learning from abundant data, which can overcome the shortages induced by handcrafted features.
Especially, after the success of long-short term memory (LSTM) networks in capturing the complex temporal dependencies \cite{ma2015long,alahi2016social}, many works have applied LSTM to vehicle trajectory prediction. 
For example, Zyner {\it et al.} \cite{zyner2018recurrent} proposed an LSTM-based model to predict a potential direction that the driver would take at an intersection.
Xin {\it et al.} \cite{xin2018intention} proposed a dual LSTM-based model to estimate driver intentions and predict future trajectories.
These approaches take the past trajectory of a target vehicle as input and can achieve high accuracy of trajectory prediction.
However, they ignore the impact of nearby vehicles on the target vehicle.

Later on, inspired by the interactive nature of drivers, researchers began to add the effects of nearby vehicles to deep-learning-based models. For example, Deo {\it et al.} \cite{deo2018multi} proposed a maneuver LSTM (M-LSTM), which takes the past trajectories of a target vehicle and its nearby vehicles as inputs.
However, this method only aggregates all trajectories together but ignores the different effects of nearby vehicles on the target one. 
To improve the M-LSTM, Deo {\it et al.} \cite{deo2018convolutional} proposed a network named CS-LSTM where a social tensor and a convolutional social pooling mechanism are introduced to model and capture the spatial interactions, respectively.
Similarly, Zhao {\it et al.} \cite{zhao2019multi} proposed a multi-agent tensor fusion (MATF) network, which introduces a spatial tensor to represent spatial relations among vehicles.
However, since both the social tensor and the spatial tensor in CS-LSTM and MATF only retain the spatial relationships at the last timestamp of past trajectories, the spatial-temporal dependencies are ignored by them.
To address this issue, Dai {\it et al.} \cite{dai2019modeling} proposed a spatio-temporal LSTM to consider the dynamic effects of six closest vehicles on the target vehicle. 
Similarly, Hou {\it et al.} \cite{hou2019interactive} proposed a structural-LSTM network to consider the dynamic effects of five nearby vehicles on the target vehicle.

Although the above deep-learning-based approaches have made great progress in improving the accuracy of vehicle trajectory prediction, there are still two limitations.
Firstly, the spatial relations of vehicles are essentially non-Euclidean, such that it is difficult to explain the physical meaning of features when using LSTM to model the spatial-temporal interactions.
Therefore, these LSTM-based approaches would be neither efficient nor intuitive in modeling the spatial-temporal interactions.
Secondly, these models taking the LSTM as the backbone network require an intensive computation power, and most of them only predict one target vehicle's future trajectory each time. 
Thus, the computation time would increase exponentially when predicting future trajectories of all neighbor vehicles, which would be not suitable for the real-time decision-making of autonomous vehicles.

Therefore, to overcome the two aforementioned limitations, this paper proposes an efficient and fast network called graph-based spatial-temporal convolutional network (GSTCN), which can simultaneously predict future trajectory distributions of all neighbor vehicles. 
Inspired by the fact that GCN can capture the spatial dependencies in a traffic network with a faster computation speed and a higher efficiency than the LSTM-based methods \cite{chai2018bike,guo2019attention,geng2019spatiotemporal,liu2019contextualized,li2019grip,mohamed2020social}, we design a spatial graph convolutional module to learn the spatial dependencies among vehicles. 
To distinguish the respective effects of neighbor vehicles on a vehicle, we propose a weighted adjacency matrix which is embedded into the spatial graph convolutional module. 
Moreover, to capture the correlations of the features between the prediction and past time horizons, we design a CNN-based temporal dependency extractor (TDE) operated in the temporal dimensions.
In this way, features in the past time horizon are mapped into the prediction time horizon for analysis of vehicle trajectory evolutions. 
In addition, the spatial-temporal features are fed into a GRU-based encoder-decoder to generate future trajectory distributions.
The main contributions of our work are as follows.

(1) The backbones of our GSTCN are GCN and CNN, so it has a smaller model size and a faster inference speed than those of LSTM-based models, making the real-time prediction of all nearby vehicles' trajectories possible. 

(2) A weighted adjacency matrix is proposed to describe the intensity of mutual influence between two vehicles, and the ablation study demonstrates the network with it has better performance than that with an unweighted adjacency matrix.

(3) The GSTCN generates the probability distributions over the future trajectories, which can describe the stochastic behaviors of the human drivers compared to models predicting deterministic trajectories, especially in a long prediction horizon. 

The rest of this paper is organized as follows. The problem description is given in Section II. We present a detailed description of the proposed network in Section III. The experimental results and analysis are given in Section IV. Finally, conclusions and possible future works are drawn in Section V.

\section{Problem Description of Vehicle Trajectory Prediction}
Vehicle trajectory prediction is an important assistant function of autonomous vehicles, which can help autonomous vehicles to assess the possibilities of risks and to plan appropriate trajectories in advance.
Similar to the work in \cite{deo2018convolutional}, this paper formulates the vehicle trajectory prediction as estimating the future trajectory distributions given past trajectories.
The difference is that we predict future positions for all neighbor vehicles simultaneously, which can provide autonomous vehicles with more detailed information about future situations.

\begin{figure}
\centerline{\includegraphics[scale=0.43]{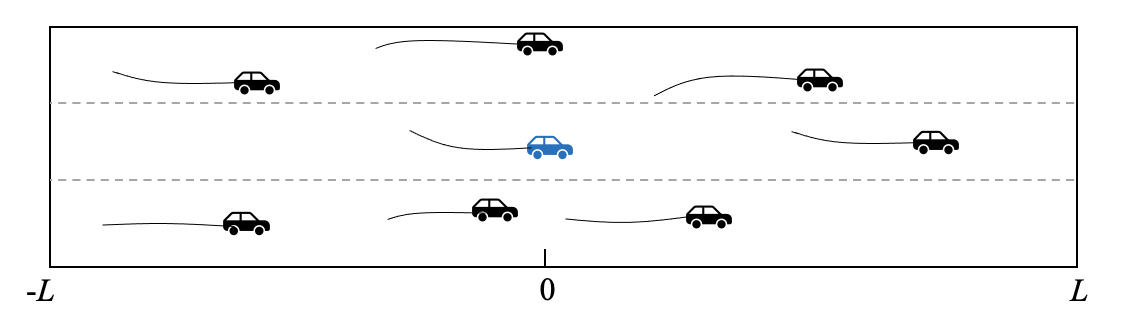}}
\caption{Illustration of a predicted scene. The blue vehicle is an autonomous vehicle, and all the black vehicles can be observed by the autonomous vehicle in the past time horizon $T$. 
The lines at the rears of each vehicle represent the past trajectories.
Without loss of generality, we predict the probability distributions over the future trajectories of all vehicles simultaneously.}
\label{scene}
\end{figure}

\begin{figure*}
\centerline{\includegraphics[scale=0.43]{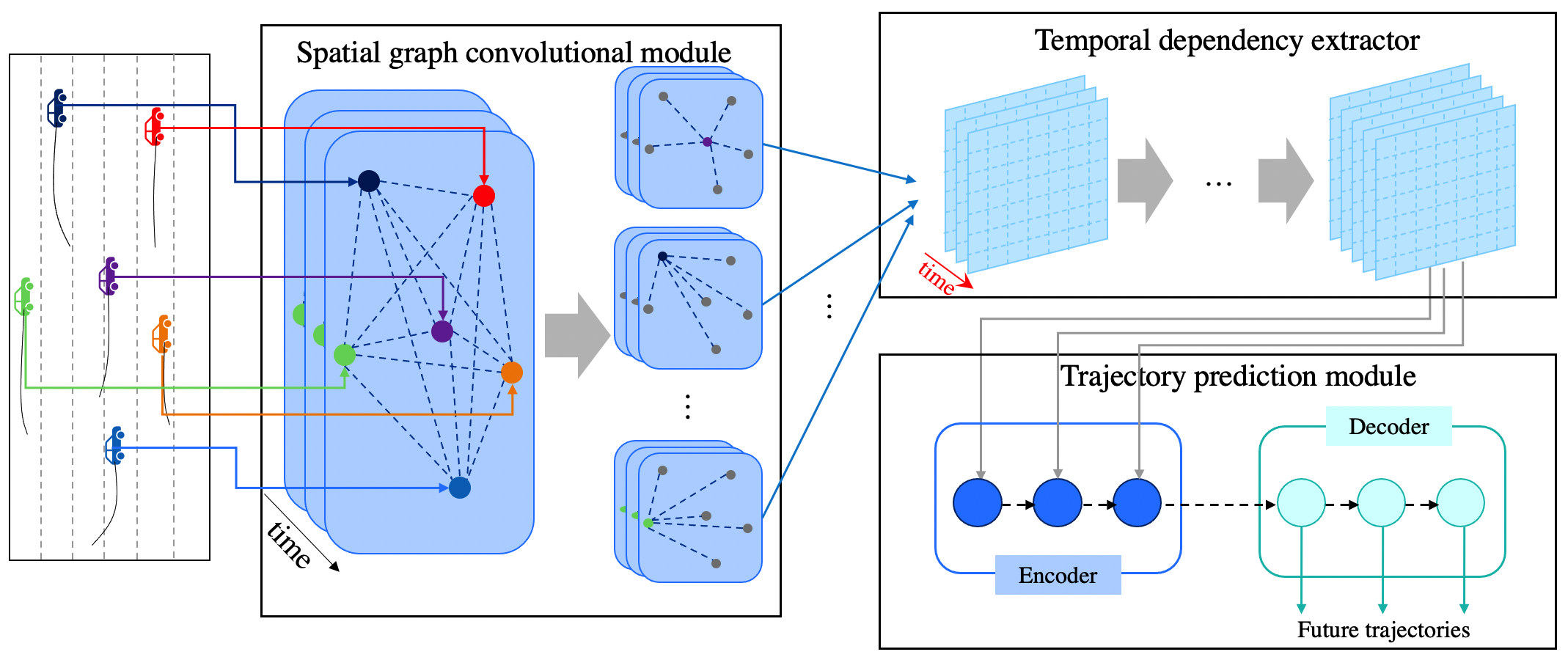}}
\caption{{Architecture of the proposed graph-based spatial-temporal convolutional network (GSTCN). The spatial graph convolutional module takes the past trajectories of vehicles as inputs and learns the spatial interdependencies among nearby vehicles, and then the temporal dependency extractor extracts features in the temporal dimensions. Finally, the spatial-temporal features are sent to the trajectory prediction module to compute the future trajectory distributions.}}
\label{pipeline}
\end{figure*}

To formulate this problem, we first introduce some notations. Vehicles' positions $X$ over a past time horizon $T$ are denoted as:
\begin{equation}
X=[P_1,P_2,\cdots,P_t,\cdots,P_{T}],
\end{equation}
where,
\begin{equation}\label{pt}
P_t=[(x_t^1,y_t^1),\cdots,(x_t^n,y_t^n),\cdots,(x_t^N,y_t^N)],
\end{equation}
are the coordinates at the time $t$, and $N$ is the number of vehicles. 
As shown in Fig. \ref{scene}, we assume that the autonomous vehicles can observe the motions of vehicles within $\pm L$ meters longitudinally and two adjacent lanes laterally, and can collect their past trajectories with a certain frequency.
The trajectory distributions in the future time horizon $F$ is denoted as:
\begin{equation}
Y = [\hat{P}_{T+1},\hat{P}_{T+2},\cdots,\hat{P}_{T+t},\cdots,\hat{P}_{T+F}],
\end{equation}
with,
\begin{equation}
\hat{P}_{T+t}=[(\hat{x}_{T+t}^1,\hat{y}_{T+t}^1),\cdots,(\hat{x}_{T+t}^n,\hat{y}_{T+t}^n),\cdots,(\hat{x}_{T+t}^N,\hat{y}_{T+t}^N)].
\end{equation}
We follow the assumption in \cite{mohamed2020social} that the predicted coordinates are random variables satisfying bi-variable Gaussian distribution, i. e., 
\begin{equation}
(\hat{x}_t^n,\hat{y}_t^n)\sim \mathcal{N}(\hat{\mu}_t^n,\hat{\sigma}_t^n,\hat{\rho}_t^n),
\end{equation}
where $\hat{\mu}_t^n$ is the mean, $\hat{\sigma}_t^n$ is the standard deviation, and $\hat{\rho}_t^n$ is the correlation.

Thus, the trajectory prediction problem can be summarized as follows.
Given all neighbor vehicles' positions $X$ over a past time horizon $T$, the aim is to predict their trajectory distributions $Y$ in the future time horizon $F$.

\section{Graph-based Spatial-Temporal Convolutional Network}\label{s-model}
To solve the trajectory prediction problem, a key challenge is to figure out how vehicles affect each other, i.e., spatial-temporal dependencies among vehicles.
Moreover, since predicted trajectories are time series, another challenge lies in how to tackle the sequence generation task.
To address the two challenges, we propose a graph-based spatial-temporal convolutional network (GSTCN) for vehicle trajectory prediction.
The overall architecture of our proposed network is shown in Fig. \ref{pipeline}, in which the spatial graph convolutional module and the TDE are used to capture the spatial-temporal dependencies, and the trajectory prediction module is applied to predict future trajectories.
We introduce each component in the following subsections.

\subsection{Spatial graph convolutional module}\label{sgcm}
\subsubsection{Generation of spatial-temporal graph using trajectories}
Inspired by the topological structure of the graph, we model the interactions among vehicles as a spatial-temporal graph.
The spatial-temporal graph is defined as $G=\{G_t|\forall t \in \{1,\cdots,T\}\}$, where $G_t$ is the spatial graph representing the spatial relations of vehicles at the time $t$.
Supposing that there are $N$ vehicles in a scene, we define the spatial graph as $G_t=\{V_t,E_t\}$, where 
$V_t=\{v_t^n|\forall n \in \{1,\cdots,N\}\}$ is the set of all vertices.
Each vertex $v_t^n$ represents an individual vehicle in the scene, and the attribute of $v_t^n$ is the coordinate $(x_t^n, y_t^n)$.
$E_t$ is the set of all edges, and each edge represents the mutual effects between vehicles.

\begin{figure}
\centerline{\includegraphics[scale=0.4]{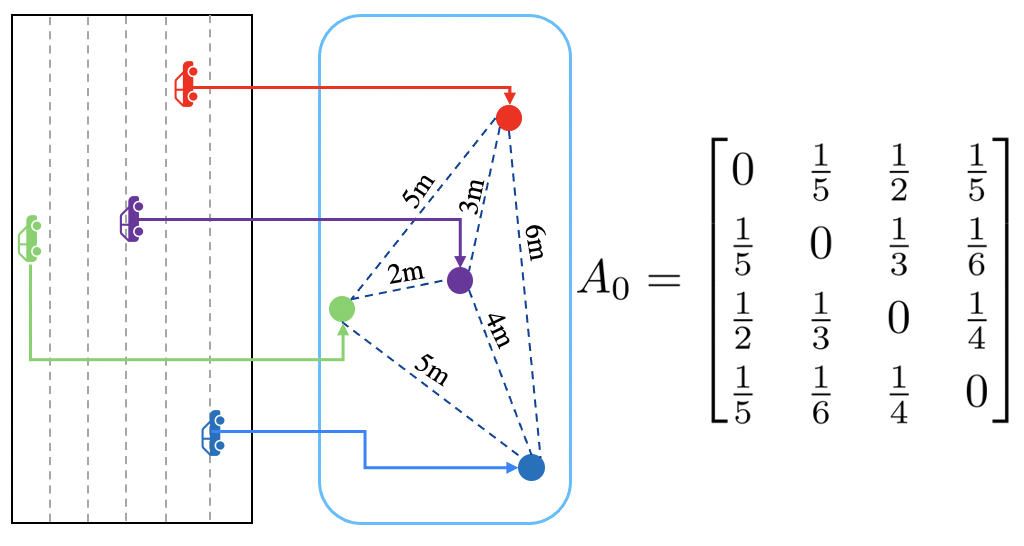}}
\caption{The example of the graph generation at one timestamp. The color of each vehicle is corresponding to the color of each vertex in the graph. The weights in the adjacency matrix $A_0$ are marked as the reciprocal of distances.}
\label{graph}
\end{figure}

Generally, a vehicle has different effects on other vehicles.
For example, the sudden deceleration of a vehicle would cause close vehicles to slow down or change lanes, but have little influence on vehicles far away from it.
Therefore, to distinguish the intensity of interactions between two vehicles, we consider that each edge in $E_t$ should be assigned to different weights.
In this work, we introduce $A_t\in \mathbb{R}^{N\times N}$ as a weighted adjacency matrix whose entries represent how strong the interactions between two vehicles could be.
Considering that two vehicles with closer distances have stronger effects on each other, we use the reciprocal of the distance to measure the weight between two vehicles so that closer vehicles have higher weights. Hence, $A_t$ can be written as,
\begin{equation}\label{am0}
A_t = \begin{pmatrix}
0 & \frac{1}{d_{1,2}} & \frac{1}{d_{1,3}} & \cdots & \frac{1}{d_{1,N}}\\ 
\frac{1}{d_{2,1}} & 0 & \frac{1}{d_{2,3}} & \cdots & \frac{1}{d_{2,N}}\\
\frac{1}{d_{3,1}} & \frac{1}{d_{3,2}} & 0 & \cdots & \frac{1}{d_{3,N}}\\
\vdots & \vdots & \vdots & \ddots & \vdots\\
\frac{1}{d_{N,1}} & \frac{1}{d_{N,2}} & \frac{1}{d_{N,3}} & \cdots & 0
\end{pmatrix},
\end{equation}
where $d_{i,j}$ in $A_t$ denotes the distance between the vehicles $i$ and $j$ at the time $t$.
A toy example about the generation of a spatial graph is given in Fig. \ref{graph}, in which each vertex in the graph is corresponding to a vehicle and the reciprocal of distances represent the weights in the weighted adjacency matrix $A_0$.
At each time $t$ in the past time horizon, we can construct a spatial graph $G_t$. 
By stacking $G_1,\cdots,G_{T}$, we get the spatial-temporal graph $G$, which is the input of our network. In the spatial-temporal graph, we define $V\in \mathbb{R}^{ C\times T\times N}$ as the stacked tensor of all $V_t$ in the past time horizon, and $C$ is set to be 2 to represent the two-dimensional coordinate $(x, y)$. Similarly, the adjacency matrix of $G$ is denoted as $A \in \mathbb{R}^{T \times N \times N}$, which is the stack of $A_1,\cdots,A_{T}$.

\subsubsection{Spatial graph convolution}
The spatial-temporal graph contains raw information about dependencies among vehicles, so we should use well-designed networks to extract these dependencies from the graph.
In the spatial dimension, a vehicle's future trajectory is highly dependent on the motions of its nearby vehicles.
To capture the spatial dependencies, existing works \cite{deo2018convolutional,zhao2019multi} use the number of zeros in the grid to represent distances between vehicles, which is inefficient.
Since GCN directly operates on the vertices of a graph and has shown its effectiveness to capture the spatial dependencies between one vertex and its neighbors \cite{yu2017spatio}, we introduce the spatial graph convolution in this work.

Extended from standard two dimensional convolution \cite{kipf2016semi}, the graph convolution operation is expressed as:
{
\begin{equation}
Z^{(l+1)} = f(\Lambda_t^{-\frac{1}{2}}\hat{A}_t\Lambda_t^{-\frac{1}{2}}Z^{(l)}\textbf{W}^{(l)}),
\end{equation}}
where $Z^{(l)}$ denotes the feature matrix of vertices in layer $l$. $f(\cdot)$ is an activation function, {$\hat{A}_t=A_t+I$, $I\in \mathbb{R}^{N\times N}$} is the identity matrix, {$\Lambda_t\in \mathbb{R}^{N\times N}$} is the diagonal node degree matrix of {$\hat{A}_t$}, $\textbf{W}^{(l)}$ is the parameters matrix of the layer $l$.
The aim of computing {$\Lambda_t^{-\frac{1}{2}}\hat{A}_t\Lambda_t^{-\frac{1}{2}}$} is to normalize the adjacency matrix, which can speed up the learning process of GCN \cite{kipf2016semi}.

\begin{figure}
\centering
\subfigure[]{
\includegraphics[width=3.5cm]{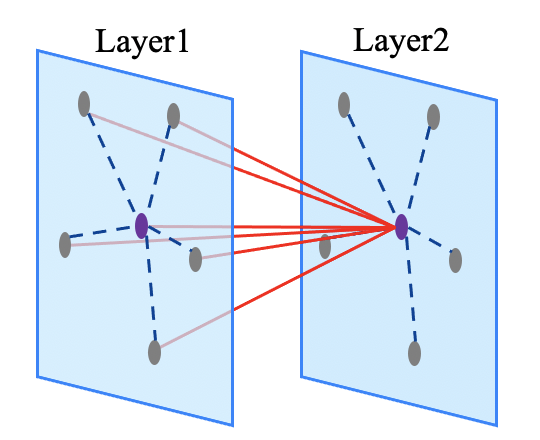}
}
\quad
\subfigure[]{
\includegraphics[width=3.5cm]{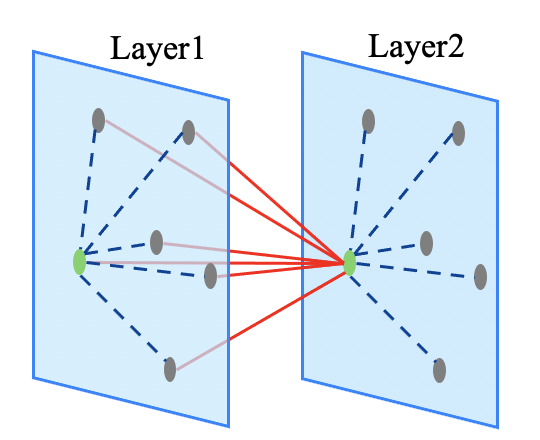}
}
\caption{Schematic plot of the graph convolution operation on each vertex, the red lines represent the weights: (a) graph convolution operation on the purple vertex; (b) graph convolution operation on the green vertex.}
\label{gcni}
\end{figure}

As shown in Fig. \ref{gcni}, the graph convolution operations apply a weighted sum to the features of a target vehicle and its surrounding vehicles, and then pass the results to the next layer.
The feature of the target vehicle is considered because the state of the target vehicle also impacts its future motions.
Note that the shapes of features are the same before and after the graph convolution.

\subsection{{Temporal dependency extractor}}

In the temporal dimension, the future motions of a vehicle are highly dependent on its own past trajectory. 
For instance, a vehicle that is conducting the maneuver of lane change is most likely to continue this maneuver in the next few seconds.
In addition, the effects of the surrounding vehicles on the target vehicle are time-varying.
To capture the temporal dependencies, most of the existing works \cite{deo2018convolutional,altche2017lstm} rely on LSTM. However, LSTM would lead to low training efficiency and slow computation speed \cite{vaswani2017attention}.
Inspired by the work in \cite{mohamed2020social}, we design a CNN-based temporal dependency extractor (TDE) to extract the temporal features. Since this module depends on convolution operations, it has a smaller parameter size and a faster inference speed than those of LSTM.

\begin{figure}
\centering
\subfigure[]{
\includegraphics[width=3.2cm]{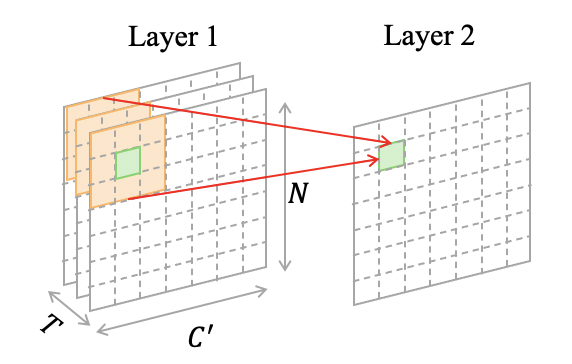}
}
\quad
\subfigure[]{
\includegraphics[width=4.4cm]{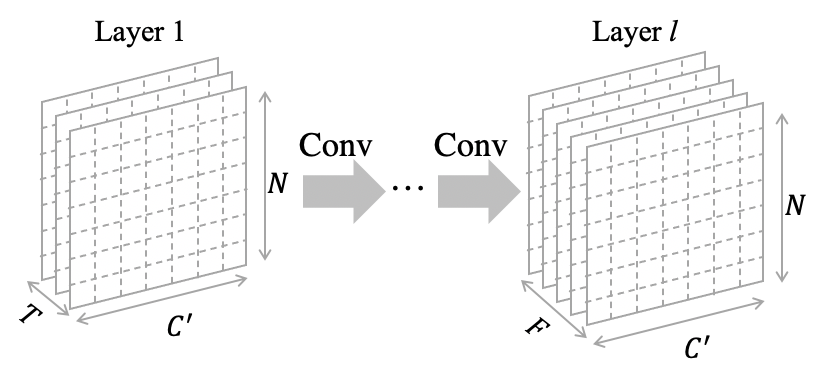}
}
\caption{Schematic plot of the temporal dependency extractor: (a) the filter integrates temporal information from different channels together; (b) input and output of the temporal dependency extractor.}
\label{cnn}
\end{figure}

The procedure of this module is as follows.
After the neighboring information of each vertex has been captured by the spatial graph convolution operations, we get a three-dimensional tensor $H \in \mathbb{R}^{C' \times T \times N}$ which contains the extracted spatial features.
Then, we generate $H' \in \mathbb{R}^{T \times C' \times N}$ from $H$ with the transposition of dimensions. Each channel of $H'$ contains the spatial features of all vehicles at the corresponding timestamp.
The TDE takes $H'$ as inputs to learn the evolving tendency of each vehicle from its own past motions and nearby vehicles' dynamic interactions.
The TDE takes the length of the past time horizon as the number of input channels.
{
As shown in Fig. \ref{cnn}(a), a filter consists of multiple kernels, and each kernel is designed to learn the interactions among vehicles at one past timestamp by operating convolutions in one past temporal feature map. We set the number of kernels in a filter to be equal to the length of the past time horizon, so a filter can integrate past temporal information together into one feature map by performing element-wise addition on its kernels’ output feature maps. Besides, we set the number of filters in TDE to be equal to the length of the prediction time horizon. In this way, the TDE can extract temporal dependencies by learning the mapping relation between the input features and future temporal features.
}
The output of the TDE is a tensor with shape $(F \times C' \times N)$, as shown in Fig. \ref{cnn}(b).

\subsection{Trajectory prediction module}
After capturing the spatial-temporal features, we should generate future trajectory distributions, which is a typical sequence generating task. 
Inspired by the excellent performance of gated recurrent unit (GRU) network on sequence-based tasks and its cheap computation cost \cite{cho2014learning}, we design the trajectory prediction module as a GRU-based encoder-decoder network.
{In our model, the spatial graph convolutional module learns the spatial dependencies among vehicles from the inputs, and then the temporal dependency extractor further extracts the temporal dependencies. Although the operations of these two modules are consecutive in the implementation steps, the spatial dependencies and temporal dependencies are not extracted at the same time, which would weaken the correlation between the spatial and temporal dependencies contained in the extracted features. Therefore, in this module, the encoder GRU is designed to strengthen the correlation between spatial dependencies and temporal dependencies, and the decoder GRU generates the probability distributions of future trajectories.}
The predicted coordinate $(\hat{x}_t^n,\hat{y}_t^n)$ is given by $(\hat{x}_t^n,\hat{y}_t^n)\sim \mathcal{N}(\hat{\mu}_t^n,\hat{\sigma}_t^n,\hat{\rho}_t^n)$.
Note that the encoder GRU or the decoder GRU used for each vehicle has shared weights, which guarantees the generalization of the model even if the number of neighbor vehicles varies.

\section{Experimental Evaluation}
\subsection{Datasets}
\begin{figure}
\centering
\subfigure[I-80.]{
\includegraphics[width=3cm]{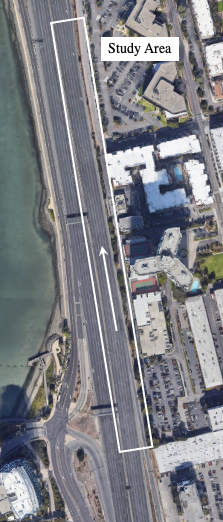}
}
\quad
\subfigure[US-101.]{
\includegraphics[width=3cm]{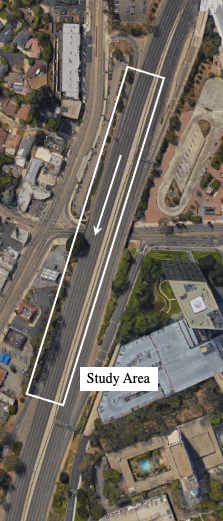}
}
\caption{Top view of the two areas to be studied. Figures are downloaded from Google Maps.}
\label{i80us101}
\end{figure}

The model is trained on two public vehicle trajectory datasets: I-80 and US-101 in NGSIM \cite{ngsim2}, in which trajectories are recorded with a frequency of 10Hz under real freeway scenarios. 
Both datasets contain vehicle trajectories in mild, moderate, and heavy traffic scenarios for 45 minutes. The rich scenarios in the datasets are suitable for the evaluation of the robustness and effectiveness of the proposed network.
The freeways of I-80 and US-101 to be studied are shown in Fig. \ref{i80us101}, where the white arrows indicate driving directions.

To make a fair comparison, we follow the same training strategy in \cite{deo2018convolutional}: the raw data are downsampled to 5Hz, and the trajectories are split into segments of 8 seconds, in which the first 3 seconds of each segment are taken as the past time horizon and the remaining 5 seconds are treated as the prediction time horizon.
We finally get 13,218 segments of trajectories, and all of them are randomly split into training, validation, and testing sets.

\subsection{Evaluation metrics}
In order to achieve quantitative metrics, the root mean square error (RMSE) of the predicted trajectory and the ground truth is used to evaluate performance.
The RMSE is calculated as,
\begin{equation}
RMSE_t = \sqrt{\frac{1}{N}\sum_{n=1}^{N}[(\hat{x}_{t}^n-x_t^n)^2+(\hat{y}_{t}^n-y_t^n)^2]}
\end{equation}
where $\hat{x}_{t}^n$ and $\hat{y}_{t}^n$ are the predicted coordinate of vehicle $n$ at time $t$. 
We report the RMSE values for different prediction time horizons (from 1 to 5 seconds).
We note when evaluating predicted probability distributions, previous works \cite{deo2018convolutional,zhao2019multi} only calculate the RMSE of the mean values and the ground truth but ignores the predicted standard deviations and correlations.  
Thus, in order to make a more comprehensive evaluation, we report the lowest error based on 5 random samplings, which is similar to \cite{mohamed2020social}.

\subsection{Implementation details}
We implement the proposed model using PyTorch as that in \cite{paszke2017automatic}.
Several implementation details are given as follows.

\subsubsection{{Data preprocessing}} {
In a real scenario with imperfect sensors which, for example, result in missing or abnormal data, it is necessary to preprocess the raw data, where it is required to remove the abnormal data and infer the missing data. We have introduced a data preprocessing module to complete these tasks. With this module, the abnormal data can be detected and removed by using one of existing anomaly detection methods, and missing data can be inferred by cubic Hermite interpolation.}

\subsubsection{Scene size} We set the autonomous vehicle's horizon of sight as follows. The autonomous vehicles can observe the motions of vehicles within the range of $\pm$ 100 meters longitudinally and two adjacent lanes laterally.

\subsubsection{Input embedding} We use a 32-channel convolutional layer with  $1\times1$ kernel size to increase the dimensions of spatial coordinates, which can improve the learning ability of the network \cite{alahi2016social}.

\subsubsection{Temporal dependency extractor} Residual connections are used in this module.
The kernel size is $3\times3$, and the padding is set to be 1 to guarantee that the shape of features is not changed after the convolutions.

\subsubsection{Trajectory prediction module} Both the encoder and decoder are one-layer GRUs.
A linear layer is used to make sure the outputs of the decoder have the expected shape, and also dropout (with 0.5 probability) is applied to prevent overfittings.

\subsubsection{Training loss}
Since the outputs of our network are the probability distributions, we train the network by minimizing the negative log-likelihood loss:
\begin{equation}
Loss=-\sum_{n=1}^{N}\sum_{t=T+1}^{T+F}{\rm log}(\mathbb{P}(x_t^n,y_t^n|\hat{\mu}_t^n,\hat{\sigma}_t^n,\hat{\rho}_t^n)),
\end{equation}
where $\mathbb{P}(x_t^n,y_t^n|\hat{\mu}_t^n,\hat{\sigma}_t^n,\hat{\rho}_t^n)$ denotes the likelihood of the ground truth position $(x_t^n,y_t^n)$ over the predicted probability distribution.

\subsubsection{Training process} The GSTCN is trained on the NVIDIA GTX1080Ti GPU.
The batch size is set to be 128, and we train the model for 250 epochs using Stochastic Gradient Descent (SGD) optimizer with an initial learning rate of 0.1. The learning rate is multiplied by 0.1 every 80 epochs to speed up the convergence of the loss.

\subsubsection{Model configuration}
\begin{figure}
\centerline{\includegraphics[scale=0.35]{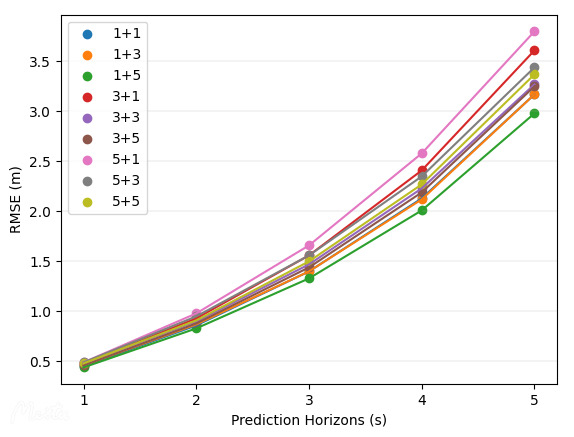}}
\caption{Comparison of RMSE for the models with different numbers of layers. The legend shows $(\alpha+\beta)$, which means that there are $\alpha$ layers in the spatial graph convolutional module and $\beta$ layers in the TDE.}
\label{hyper}
\end{figure}

We choose appropriate hyperparameters to improve the performance of our network. We first test how the numbers of layers in the spatial graph convolutional module and the TDE affect the performance. As shown in Fig. \ref{hyper}, the best model has one layer in the spatial graph convolutional module and five layers in the TDE. Then, we fix the layer number and vary the number of hidden units in GRU as the number from $\{16, 32, 64, 100, 128\}$. 
As shown in Table \ref{gru}, the model has the lowest RMSE values when the number of hidden units is set to be 32.

\begin{table}
\caption{Comparison of RMSE for the models with different numbers of hidden units in GRU. Data are converted into the meter unit.}
\label{gru}
\centering
\begin{center}
\begin{tabular}{p{50pt}<{\centering}||p{20pt}<{\centering}p{20pt}<{\centering}p{20pt}<{\centering}p{20pt}<{\centering}p{20pt}<{\centering}}
\hline
Prediction &\multirow{2}{*}{16} & \multirow{2}{*}{32} & \multirow{2}{*}{64} & \multirow{2}{*}{100} & \multirow{2}{*}{128}\\
Horizon(s) &  & & & &   \\
\hline
\hline
1 & 0.48 & 0.44  & 0.43 & 0.42 & \textbf{0.42} \\[3pt]
2 & 0.87 & \textbf{0.83}  & 0.84 & 0.83 & 0.84 \\[3pt]
3 & 1.36 & \textbf{1.33}  & 1.36 & 1.37 & 1.36 \\[3pt]
4 & 2.05 & \textbf{2.01}  & 2.05 & 2.05 & 2.03 \\[3pt]
5 & 3.06 & \textbf{2.98}  & 3.03 & 3.03 & 2.99 \\[3pt]
\hline
\hline
Average & 1.56 & \textbf{1.52}  & 1.54 & 1.54 & 1.53 \\[3pt]
\hline
\end{tabular}
\end{center}
\end{table}

\subsection{Ablation study}
In this subsection, we conduct several ablative studies to verify the effectiveness of our scheme.

\subsubsection{{Effectiveness of different modules}}
{
Our full model mainly consists of three modules. To verify the effectiveness of each module on vehicle trajectory prediction, we train three variants of our model with different modules: GSTCN without GCN, GSTCN without TDE and GSTCN without GRU. As shown in Table \ref{ablation0}, the removal of any module in GSTCN will cause an increase in RMSE, which indicates the effectiveness of each module.
When removing GCN or TDE from our full model, the features sent to the GRU lack the extracted spatial dependencies or temporal dependencies. As a result, the GRU fails to integrate the correlation between the spatial dependencies and temporal dependencies and cannot reasonably generate future trajectories.
If the GRU is removed, the weak correlation between the spatial and temporal dependencies cannot be strengthened.
Only after both the spatial dependencies and temporal dependencies are sent to the trajectory prediction module, can our model achieve the best performance, which indicates that the GCN and TDE are complementary in our full model, and the GRU encoder-decoder can strengthen the correlation between spatial dependencies and temporal dependencies from the extracted features and generate more accurate future trajectories.
}

\begin{table}
\caption{{Comparison of RMSE for the variants of GSTCN. Data are converted into the meter unit.}}
\label{ablation0}
\centering
\begin{center}
\begin{tabular}{p{40pt}<{\centering}||p{40pt}<{\centering}p{40pt}<{\centering}p{40pt}<{\centering}p{20pt}<{\centering}}
\hline
Prediction & GSTCN & GSTCN & GSTCN & \multirow{2}{*}{GSTCN}\\
Horizon(s) & W/O GCN & W/O TDE & W/O GRU &   \\
\hline
\hline
1 & 0.66 & 0.62 & 0.82 & \textbf{0.44} \\[3pt]
2 & 1.64 & 1.53 & 1.58 & \textbf{0.83} \\[3pt]
3 & 2.73 & 2.61 & 2.50 & \textbf{1.33} \\[3pt]
4 & 3.98 & 3.84 & 3.73 & \textbf{2.01} \\[3pt]
5 & 5.30 & 5.22 & 5.29 & \textbf{2.98} \\[3pt]
\hline
\hline
Average &  2.86 & 2.76 & 2.78 & \textbf{1.52}  \\[3pt]
\hline
\end{tabular}
\end{center}
\end{table}

\subsubsection{Influence of different locations}
\begin{figure}
\centerline{\includegraphics[scale=0.35]{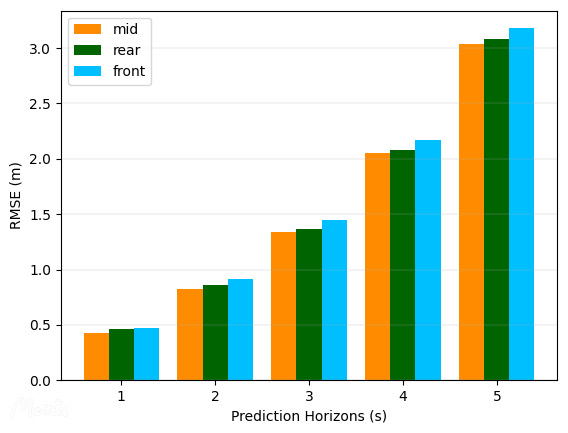}}
\caption{Comparison of RMSE for vehicles in different locations of a scene: middle, rear and front.}
\label{rmse}
\end{figure}

The proposed network can simultaneously predict all neighbor vehicles' future trajectories based on their past trajectories.
Therefore, it is necessary to analyze the prediction errors of vehicles at different locations. As shown in Fig. \ref{rmse}, we report the RMSE values for vehicles located in the middle, front, and rear of the scene.

We note that vehicles located in the middle of the scene have the lowest RMSE values for all prediction horizons, which is consistent with the intuition that vehicles located in the middle have more neighboring information as inputs than vehicles at the edge.
An interesting phenomenon is that the prediction errors of vehicles located in the rear of the scene are always lower than those of the vehicles in the front of the scene. 
This is because drivers are more likely to be affected by the vehicles in front of them according to the driving experience.
However, when predicting the trajectories of the vehicles in the front of the scene, we can only utilize the motion information of the vehicles behind them.
Thus, the predictions of the GSTCN are consistent with intuition.

\subsubsection{GSTCN with different weighted adjacency matrices}
The weights in the adjacency matrix measure the intensities of interactions among vehicles, so we should appropriately mark the weights.
In Section \ref{sgcm}, we have introduced the reciprocal of the distance between two vehicles to mark the weights based on prior knowledge. This means that the closer the distance between vehicles the stronger the mutual effects are. 
Therefore, to validate this prior knowledge has a positive influence on the performance of our model, we introduce two other ways to measure the weights: (1) we directly use the distances between two vehicles to mark the weights; (2) similar to existing works \cite{li2019grip,kipf2016semi}, we set all elements in the adjacency matrix to ones as a baseline.

\begin{table}
\caption{Comparison of RMSE for the models with different weighted adjacency matrices. Data are converted into the meter unit.}
\label{ablation}
\centering
\begin{center}
\begin{tabular}{p{50pt}<{\centering}||p{40pt}<{\centering}p{35pt}<{\centering}||p{50pt}<{\centering}}
\hline
Prediction &  \multirow{2}[2]{*}{Distance} & \multirow{2}[2]{*}{Ones} & Reciprocal of\\
Horizon(s) &  & & distance \\
\hline
\hline
1 &  0.48 & 1.10 & \textbf{0.44}\\[3pt]
2 &  0.88 & 2.04 & \textbf{0.83}\\[3pt]
3 &  1.41 & 3.09 & \textbf{1.33}\\[3pt]
4 &  2.10 & 4.29 & \textbf{2.01}\\[3pt]
5 &  3.08 & 5.70 & \textbf{2.98}\\[3pt]
\hline
\hline
Average &  1.59 & 3.24 & \textbf{1.52}\\[3pt]
\hline
\end{tabular}
\end{center}
\end{table}

The RMSE values of different weighted adjacency matrices are listed in Table \ref{ablation}. 
The model with the weighted adjacency matrices defined by the reciprocal of the distance outperforms others. Therefore, the reasonable usage of the prior knowledge can help improve the performance of the model.
Note that all weighted adjacency matrices outperform the baseline, which demonstrates the effectiveness of weighted adjacency matrices.
An interesting result is that although it is contrary to the intuition when marking the weights directly using the distances, the prediction errors are still lower than that of the baseline, which is partly due to the reverse learning ability of the deep-learning-based models.

\begin{table*}
\caption{Comparison of RMSE for models that only predict trajectory of one vehicle each time. Data are in meters.}
\label{table_example}
\centering
\begin{center}
\begin{tabular}{p{50pt}<{\centering}||p{40pt}<{\centering}p{40pt}<{\centering}p{50pt}<{\centering}p{50pt}<{\centering}p{50pt}<{\centering}p{30pt}<{\centering}||p{50pt}<{\centering}}
\hline
Prediction &\multirow{2}[2]{*}{CV} & \multirow{2}[2]{*}{V-LSTM} & \multirow{2}[2]{*}{C-VGMM+VIM} & \multirow{2}[2]{*}{CS-LSTM-M} & \multirow{2}[2]{*}{CS-LSTM} & \multirow{2}[2]{*}{GRIP} & \multirow{2}[2]{*}{GSTCN-ONE}\\
Horizon(s) &  &  &  &  &   &  &  \\[3pt]
\hline
\hline 
1 & 0.73 & 0.66  & 0.66 & 1.25 & 1.03 & \textbf{0.37} & 0.42\\[3pt]
2 & 1.78 & 1.62  & 1.56 & 1.24 & 1.13 & 0.86  & \textbf{0.81}\\[3pt]
3 & 3.13 & 2.94  & 2.75 & 1.71 & 1.61 & 1.45  & \textbf{1.29}\\[3pt]
4 & 4.78 & 4.63  & 4.24 & 2.43 & 2.31 & 2.21  & \textbf{1.97}\\[3pt]
5 & 6.68 & 6.63  & 5.99 & 3.38 & 3.21 & 3.16  & \textbf{2.95}\\[3pt]
\hline
\hline
Average & 3.42 & 3.30  & 3.04 & 2.00 & 1.86 & 1.61  & \textbf{1.49}\\[3pt]
\hline
\end{tabular}
\end{center}
\end{table*}

\subsection{Baselines}
We evaluate the performance of our proposed method by comparing it with the following baselines:
\begin{itemize}
\item Constant Velocity (denoted as CV) \cite{deo2018convolutional}: This baseline uses a constant velocity Kalman filter to predict the deterministic trajectory of one vehicle. The effects of neighbor vehicles on the target vehicle are ignored.
\item Vanilla LSTM (denoted as V-LSTM) \cite{zhao2019multi}: This baseline is a simple LSTM encoder-decoder that takes the past trajectory of the target vehicle as inputs and generates a deterministic future trajectory of the target vehicle. The effects of surrounding vehicles on the target vehicle are ignored.
\item Class variational Gaussian mixture models with vehicle interaction module (denoted as C-VGMM+VIM) \cite{deo2018would}: This model uses hidden Markov models to estimate maneuver intentions and then applies variational Gaussian mixture models for the future trajectory prediction.
\item CS-LSTM with maneuvers (denoted as CS-LSTM-M) \cite{deo2018convolutional}: This LSTM-based model applies convolutional social pooling layers to tackling the spatial interactions and predicts the multi-modal trajectory distributions of the target vehicle based on maneuvers.
\item CS-LSTM \cite{deo2018convolutional}: This baseline is the same as CS-LSTM-M except that it generates unimodal prediction distribution.
\item MATF \cite{zhao2019multi}: This baseline takes past trajectories of all vehicles and the scene image of the predicted area as inputs and uses LSTM to predict deterministic trajectories of all vehicles in a scene. 
\item Graph-based interaction-aware trajectory prediction model (denoted as GRIP-ALL) \cite{li2019grip}: This baseline uses the graph to model the interactions among vehicles and uses LSTM to predict deterministic trajectories of all vehicles in a scene.
\item GRIP \cite{li2019grip}: This baseline is the same as GRIP-ALL except that it only predicts the trajectory of one vehicle in the central location of the scene. 
\end{itemize}

\subsection{Quantitative analysis for prediction results}\label{quantitative}
In this subsection, we compare the prediction errors, model sizes, and inference speeds of our method with those of the above baselines.
Since some methods can predict trajectories of all neighbor vehicles simultaneously, while others only predict the trajectory of one vehicle each time, we compare the results respectively for the sake of fairness.

\subsubsection{Performance when predicting one vehicle}
In order to compare with the methods that only predict one vehicle's future trajectory each time, we report the RMSE values of the vehicle located in the middle of the scene, as shown in the column of “GSTCN-ONE” in Table \ref{table_example}.
It shows that the GSTCN-ONE achieves the lowest prediction errors for almost all prediction horizons, demonstrating the powerful ability of our GSTCN in capturing the spatial-temporal dependencies and inferring the future trajectories.

In Table \ref{table_example}, we can see that in the prediction horizon of one second, the previous state-of-the-art model GRIP has better performance than the GSTCN. However, the deterministic trajectories generated by the GRIP fail to describe the stochastic behaviors of human drivers, so our GSTCN outperforms the GRIP in long prediction horizons, and the average RMSE values are 7.45\% lower than that of the GRIP.
Both the GSTCN and the CS-LSTM predict the probability distributions over the future trajectories, but GSTCN outperforms CS-LSTM in all prediction horizons, which demonstrates that the GCN is more effective to capture the spatial dependencies than LSTM.
In addition, we note that almost all deep-learning-based approaches have better performance than traditional and machine-learning-based methods (CV and C-VGMM+VIM).
Among all deep-learning-based methods, the ones that use the information of surrounding vehicles outperform the ones that do not (i.e., V-LSTM).
Therefore, it is crucial to consider the neighboring information when predicting the trajectories.

\subsubsection{Performance when predicting all vehicles}
Among all baselines, only the MATF and the GRIP-ALL can simultaneously predict trajectories of all vehicles.
Therefore, this paper compares the performance of the two methods with our GSTCN in the case of predicting trajectories of all vehicles.
As shown in Table \ref{table1}, our network improves the prediction accuracies for all prediction horizons.
For example, the GSTCN achieves 22.4\% average accuracy improvement compared with the GRIP-ALL.
In addition, although our model does not take the scene images as the additional inputs, it still outperforms the MATF that does.
This is because MATF only considers the spatial relationships at one timestamp but ignores the spatial-temporal dependencies, which demonstrates that capturing the spatial-temporal dependencies is far more important than processing the scene images.

\begin{table}
\caption{Comparison of RMSE for models that predict trajectories of all vehicles each time. Data are in meters.}
\label{table1}
\centering
\begin{center}
\begin{tabular}{p{40pt}<{\centering}||p{40pt}<{\centering}p{50pt}<{\centering}||p{50pt}<{\centering}}
\hline
Prediction & \multirow{2}[2]{*}{MATF} & \multirow{2}[2]{*}{GRIP-ALL} & \multirow{2}[2]{*}{GSTCN}\\
Horizon(s) &  &  &  \\[3pt]
\hline
\hline
1 & 0.67 & 0.64 & \textbf{0.44}\\[3pt]
2 & 1.51 & 1.13 & \textbf{0.83}\\[3pt]
3 & 2.51 & 1.80 & \textbf{1.33}\\[3pt]
4 & 3.71 & 2.62 & \textbf{2.01}\\[3pt]
5 & 5.12 & 3.60 & \textbf{2.98}\\[3pt]
\hline
\hline
Average & 2.70 & 1.96 & \textbf{1.52}\\[3pt]
\hline
\end{tabular}
\end{center}
\end{table}

\subsubsection{Comparison of model size and inference speed}
Model size and inference speed are two important performance factors to decide whether an algorithm can be deployed to autonomous vehicles.
The small model size guarantees that the method can work well with limited hardware resources.
The fast inference speed can guarantee that autonomous vehicles have enough time to make decisions based on predictions.
{
Since only the MATF and the GRIP-ALL can simultaneously predict trajectories of all vehicles among all baselines, we compare model sizes and inference speeds of them, as listed in Table \ref{table2}. 
}

{
The GRIP-ALL was previously the smallest model with 496.3K parameters. The model size of GSTCN is only about one tenth of that of the GRIP-ALL.
To make a fair comparison, the inference speeds of these models are tested on the NVIDIA GTX1080Ti GPU, and we compare the average time required for each model to predict the trajectory of one vehicle.
These three model simultaneously predict trajectories of 120 vehicles each time.
Our GSTCN spends an average of 0.044 ms to predict one vehicle's trajectory, which is about 7.3 times faster than the previously fastest model (GRIP-ALL).}
We achieve these improvements because the backbones of the GSTCN are GCN and CNN, which overcome the limitations induced by the recurrent architecture of LSTM.

\begin{table}
\caption{Comparison of the model sizes and inference speeds of different models. Inference time is in milliseconds.}
\label{table2}
\begin{center}
\begin{tabular}{p{50pt}||p{75pt}|p{75pt}}
\hline
\multirow{2}[2]{*}{Model} & Parameters & Average inference time\\
 & count & for one vehicle (ms)\\[3pt]
\hline
\hline
{MATF} & {15.3M (313$\times$)}& {0.370 (8.4$\times$)}\\[3pt]
\hline
GRIP-ALL & 496.3K (10$\times$) & 0.322 (7.3$\times$)\\[3pt]
\hline
\hline
GSTCN & \textbf{48.9K} & \textbf{0.044}\\[3pt]
\hline
\end{tabular}
\end{center}
\end{table}

\begin{figure*}
\centering
\subfigure[]{
\includegraphics[width=8cm]{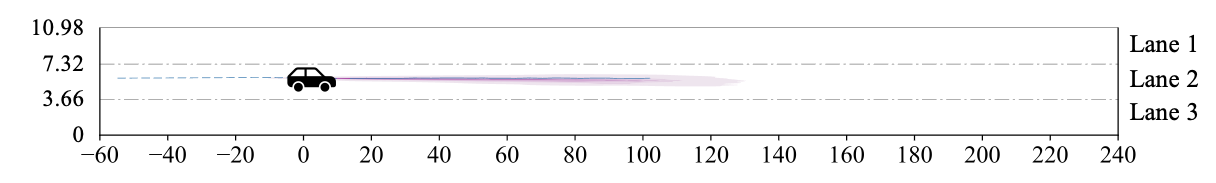}
}
\quad
\subfigure[]{
\includegraphics[width=8cm]{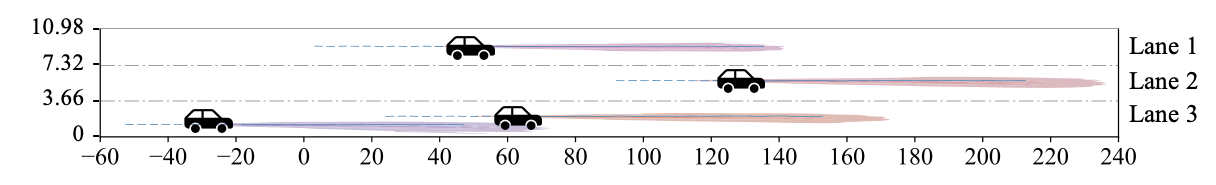}
}
\quad
\subfigure[]{
\includegraphics[width=8cm]{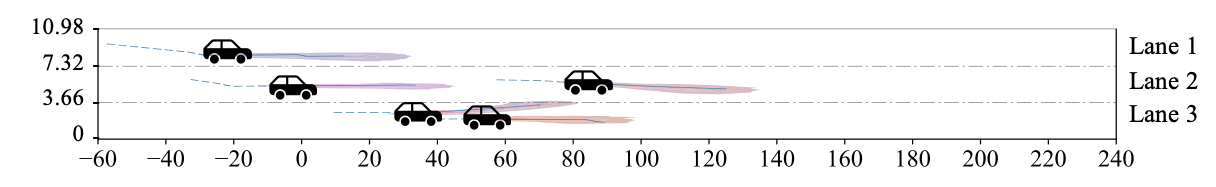}
}
\quad
\subfigure[]{
\includegraphics[width=8cm]{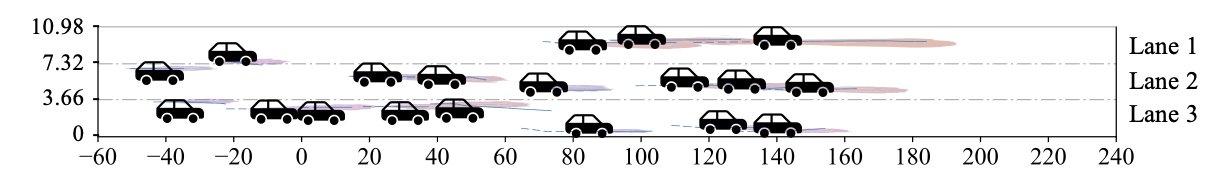}
}
\caption{Visualization of predicted trajectories under mild, moderate and congested traffic scenarios. Coordinates on the horizontal axis and the vertical axis are in meters: (a) only one vehicle; (b) four vehicles with few interactions; (c) five vehicles with some interactions in which a vehicle would conduct lane changing; (d) heavy traffic.}
\label{visual}
\end{figure*}

\subsection{Qualitative analysis for GSTCN}
In this subsection, we qualitatively analyze the prediction performance of our GSTCN by visualizing several representative predicted trajectories under mild, moderate, and heavy traffic scenarios.
All the results are sampled from the I-80 and US-101 datasets.
As shown in Fig. \ref{visual}, our model observes the past trajectories of all vehicles in the scene for 3 seconds, illustrated as dashed lines.
Then the probability distributions over all trajectories for the next 5 seconds are predicted, which are shown as color densities.
The solid lines represent the ground truth trajectories.
Overall, the predicted trajectory distributions can capture the pattern of ground truth trajectories well.

As shown in Fig. \ref{visual}(a) and (b), when the traffic condition is mild, vehicles have few interactions and tend to drive at a high speed, and the predicted distributions well prove that our network has learned this feature.
In the moderate traffic scenario, vehicles are more likely to change lanes to maximize their speeds. 
For example, as shown in Fig. \ref{visual}(c), since the front vehicle on Lane 3 has a relatively slow speed, the vehicle behind it would conduct lane changing. 
The GSTCN can capture this kind of spatial-temporal dependencies and successfully predict future distributions.
While in heavy traffic, the motions of vehicles become more complicated.
For example, as shown in Fig. \ref{visual}(d), vehicles on Lane 1 drive at relatively high speeds, while vehicles on other lanes move slowly. Our GSTCN still successfully predicts the future distributions of all vehicles, which demonstrates the robustness of our network. 

\subsection{{Robustness to imperfect data}}
{
In the previous experiments, since the NGSIM datasets have been preprocessed by the data provider, we assume that the trajectory data received by the autonomous vehicle are perfect. However, in practical application, such perfection is generally impossible to achieve. 
Therefore, we will discuss the robustness of the proposed model in the case of imperfect data in this subsection.
}

\begin{table}
\caption{{Comparison of RMSE for GSTCN with imperfect and perfect data. Data are in meters.}}
\label{table6}
\centering
\begin{center}
\begin{tabular}{p{30pt}<{\centering}||p{60pt}<{\centering}p{60pt}<{\centering}||p{30pt}<{\centering}}
\hline
Prediction & \multirow{2}[2]{*}{Case I ($\Delta$Perfect)}  & \multirow{2}[2]{*}{Case II ($\Delta$Perfect)} & \multirow{2}[2]{*}{Perfect}\\
Horizon(s) &  &  &  \\[3pt]
\hline
\hline
1 & 0.48 (+0.04) & 0.48 (+0.04) & \textbf{0.44}\\[3pt]
2 & 0.90 (+0.07) & 0.93 (+0.10) & \textbf{0.83}\\[3pt]
3 & 1.44 (+0.11) & 1.55 (+0.22) & \textbf{1.33}\\[3pt]
4 & 2.14 (+0.13) & 2.37 (+0.36) & \textbf{2.01}\\[3pt]
5 & 3.12 (+0.14) & 3.53 (+0.55) & \textbf{2.98}\\[3pt]
\hline
\hline
Average & 1.62 (+0.10) & 1.77 (+0.25) & \textbf{1.52}\\[3pt]
\hline
\end{tabular}
\end{center}
\end{table}

\subsubsection{{Case I: partially missing}}
{
In a general case, the data collected by the sensor have some missing points, so we randomly select half of the sequences from the testing set, from which randomly delete 20\% of the data points.
Before these imperfect data are fed into our model, we use cubic Hermite interpolation to infer the value of missing points.
}

\subsubsection{{Case II: totally missing}}
{
An extreme case is that a vehicle is totally undetected during a past time horizon, so it is impossible to infer reasonable data for the model. For this case, we randomly select a vehicle from the input and completely discard its trajectory data.
}

{
As shown in Table \ref{table6}, we compare the RMSE for our model with imperfect and perfect data. We can see that the RMSE for imperfect data are greater than these for the perfect data, but the increase is acceptable, which demonstrates the robustness of our model and its potential for practical applications.
In addition, when one nearby vehicle is totally undetected during the past time horizon, the RMSE increments in the long prediction horizon (3-5s) are much larger than these in case I, which indicates that the complete undetection of one vehicle has a greater impact on the accuracy of vehicle trajectory prediction than the partial undetection of several vehicles.
}

\section{Conclusions and Future Work}
In this paper, we have presented a graph-based spatial-temporal convolutional network (GSTCN) that can predict the future trajectory distributions of all vehicles in a scene simultaneously.
In our method, a weighted adjacency matrix has been proposed to distinguish different effects from nearby vehicles on the target vehicle. Based on this adjacency matrix, a spatial graph convolutional module can be used to learn the spatial dependencies among vehicles. 
The experimental results have shown that our GSTCN outperforms the main previous methods, and the small model size and fast inference speed of our GSTCN have demonstrated it has the potential to be deployed to autonomous vehicles.

In the future, several works can be done to further extend our proposed network.
Firstly, in addition to past trajectories of surrounding vehicles, there are many other supplementary data (e.g., visual scene images, high-resolution maps, and vehicle-to-vehicle communication information) that can be detected by autonomous vehicles.
Therefore, how to utilize these supplementary data to further improve the performance is worth studying.
Secondly, the effects of nearby vehicles in different directions on the target vehicle are slightly different, even if the distances are the same. Hence, we can study how to consider the effects of different directions when constructing the weighted adjacency matrix.
Finally, the aim of trajectory prediction is to enable autonomous vehicles to gain the ability to make optimal decisions.
Thus, better motion planning algorithms for autonomous vehicles based on the predictions of our GSTCN can be studied to improve traffic efficiency and security.

\appendix
{In order to better understand the predictions in Fig. \ref{visual}, we present the speeds of each vehicle in the past time horizon in Table \ref{table_appendix}.}

\begin{table}
\caption{Speeds of each vehicle in the past time horizon in Fig. \ref{visual}. $v_h$ donates the horizontal speed, and $v_v$ is the vertical speed. Vehicles in each lane are numbered from left to right. Data are in meters per second.}
\label{table_appendix}
\centering
\begin{center}
\begin{tabular}{p{10pt}<{\centering}p{30pt}<{\centering}
p{18pt}<{\centering}p{18pt}<{\centering}p{18pt}<{\centering}
p{18pt}<{\centering}p{18pt}<{\centering}p{18pt}<{\centering}}
\hline
\multirow{3}[1]{*}{Scene} & \multirow{3}[1]{*}{Samples} & \multicolumn{6}{c}{Timestamp} \\
 & & \multicolumn{2}{c}{1} & \multicolumn{2}{c}{2} & \multicolumn{2}{c}{3} \\[1pt]\cmidrule(r){3-4}\cmidrule(r){5-6}\cmidrule(r){7-8}
  &  & $v_h$ & $v_v$ & $v_h$ & $v_v$& $v_h$ & $v_v$ \\[3pt]
\hline
a & Lane 2-1 & 22.9 & 0.03 & 22.6 & 0.04 & 23.4 & -0.10 \\[3pt]
\hline
\multirow{4}[2]{*}{b} & Lane 1-1 & 16.8 & -0.03 & 14.7 & 0.02 & 15.3 & -0.22 \\[3pt]
  & Lane 2-1 & 13.3 &  0.04 & 13.7 & 0.20 & 15.0 & -0.20 \\[3pt]
  & Lane 3-1 & 11.8 & 0.18 & 11.8 & -0.12 & 11.8 & -0.06 \\[3pt]
  & Lane 3-2 & 16.0 & 0.02 & 15.5 & 0.16 & 15.4 & -0.01 \\[3pt]
\hline
\multirow{5}[3]{*}{c} & Lane 1-1 & 14.92 & -0.47 & 12.60 & -0.78 & 10.23 & 0.01 \\[3pt]
  & Lane 2-1 & 10.03 & -0.82 & 8.10 & -0.01 & 9.14 & 0.02 \\[3pt]
  & Lane 2-2 & 9.30 & -0.08 & 7.79 & -0.18 & 8.05 & -0.41 \\[3pt]
  & Lane 3-1 & 8.34 & 0.00 & 7.98 & 0.05 & 8.12 & 0.32 \\[3pt]
  & Lane 3-2 & 5.13 & 0.01 & 5.02 & 0.00 & 5.16 & 0.02 \\[3pt]
\hline
\multirow{19}[0]{*}{d} & Lane 1-1 & 1.27 & -0.06 & 1.27 & -0.06 & 1.81 & -0.06 \\[3pt]
  & Lane 1-2 & 3.23 & -0.07 & 3.30 & -0.07 & 3.44 & 0.00 \\[3pt]
  & Lane 1-3 & 3.36 & 0.01 & 3.29 & 0.01 & 3.91 & 0.00 \\[3pt]
  & Lane 1-4 & 6.06 & 0.02 & 6.71 & 0.01 & 7.19 & 0.02 \\[3pt]
  & Lane 2-1 & 0.00 & 0.00 & 1.40 & 0.00 & 3.21 & 0.00 \\[3pt]
  & Lane 2-2 & 0.64 & 0.00 & 1.06 & 0.00 & 1.38 & 0.00 \\[3pt]
  & Lane 2-3 & 3.48 & 0.00 & 2.42 & 0.00 & 3.08 & 0.00 \\[3pt]
  & Lane 2-4 & 1.72 & -0.07 & 1.70 & -0.01 & 1.71 & 0.03 \\[3pt]
  & Lane 2-5 & 3.36 & 0.00 & 4.42 & 0.00 & 6.07 & 0.01 \\[3pt]
  & Lane 2-6 & 4.41 & 0.04 & 4.80 & 0.05 & 3.77 & -0.06 \\[3pt]
  & Lane 2-7 & 4.36 & 0.05 & 3.78 & 0.00 & 3.80 & 0.00 \\[3pt]
  & Lane 3-1 & 0.16 & 0.00 & 3.12 & 0.00 & 3.03 & 0.00 \\[3pt]
  & Lane 3-2 & 3.35 & 0.00 & 2.38 & 0.00 & 1.25 & 0.00 \\[3pt]
  & Lane 3-3 & 1.56 & 0.00 & 1.48 & 0.00 & 1.52 & 0.00 \\[3pt]
  & Lane 3-4 & 3.15 & 0.00 & 3.23 & 0.00 & 3.12 & 0.00 \\[3pt]
  & Lane 3-5 & 2.01 & 0.00 & 3.79 & 0.00 & 3.28 & 0.00 \\[3pt]
  & Lane 3-6 & 6.19 & -0.24 & 6.71 & 0.00 & 4.27 & 0.00 \\[3pt]
  & Lane 3-7 & 4.47 & -0.10 & 4.37 & -0.09 & 3.92 & 0.00 \\[3pt]
  & Lane 3-8 & 3.35 & 0.01 & 3.43 & 0.01 & 3.50 & 0.00 \\[3pt]
\hline
\end{tabular}
\end{center}
\end{table}



\ifCLASSOPTIONcaptionsoff
  \newpage
\fi

%
{\small
\bibliographystyle{IEEEtran}
\bibliography{IEEEabrv,egbib}
}

%






\begin{IEEEbiography}[{\includegraphics[width=1in,height=1.25in,clip,keepaspectratio]{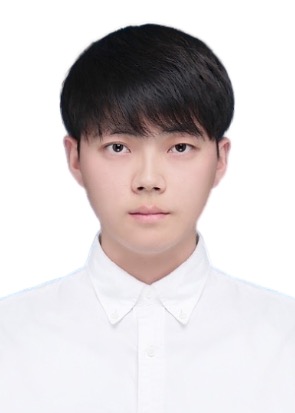}}]{Zihao Sheng}
received the B.S. degree in automation from Xi’an Jiao Tong University, Xi’an, P.R. China, in 2019, and is currently pursuing the M.S. degree in control engineering from Shanghai Jiao Tong University, Shanghai, P.R. China.
His research interests include vehicle trajectory prediction, control in autonomous driving and intelligent transportation systems.
\end{IEEEbiography}

\begin{IEEEbiography}[{\includegraphics[width=1in,height=1.25in,clip,keepaspectratio]{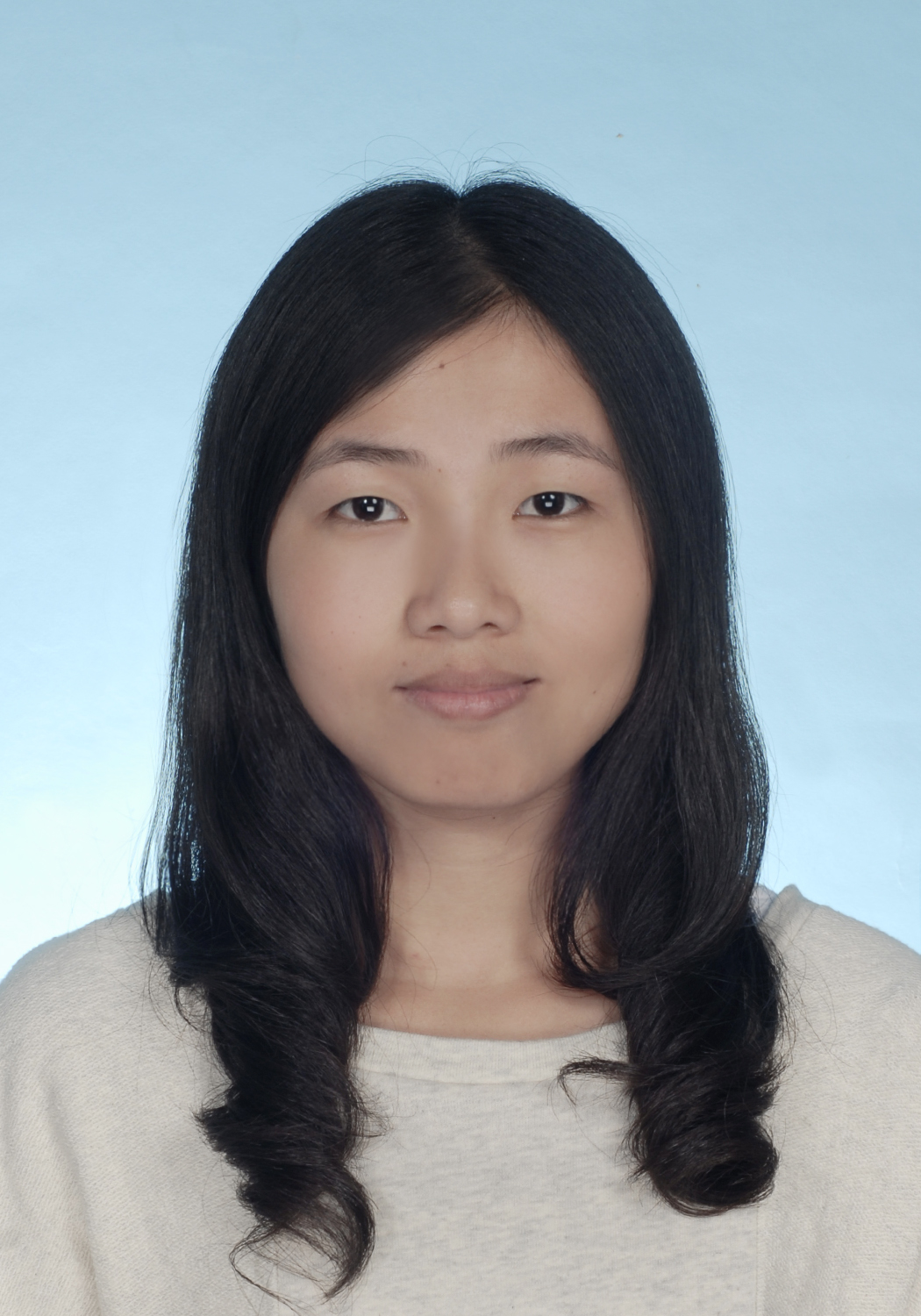}}]{Yunwen Xu}
(Member, IEEE) received the B.S. degree in automation from the Nanjing University of Science and Technology, Nanjing, P.R. China, in 2012, and the M.S. and Ph.D. degrees in control science and engineering from Shanghai Jiao Tong University, Shanghai, P.R. China, in 2014 and 2019, respectively.

She is currently a Postdoctoral Researcher with the Department of Automation, Shanghai Jiao Tong University. Her research interests include model predictive control, urban traffic modeling, and intelligent control of complex systems.
\end{IEEEbiography}

\begin{IEEEbiography}[{\includegraphics[width=1in,height=1.25in,clip,keepaspectratio]{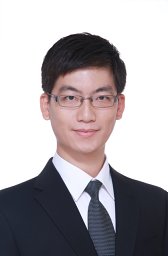}}]{Shibei Xue}
(Senior Member, IEEE) received the Ph.D. degree in control science and engineering from Tsinghua University, Beijing, P.R. China, in 2013.

He is currently an Associate Professor with the Department of Automation, Shanghai Jiao
Tong University, Shanghai, P.R. China. From 2014 to 2016, he was a Postdoctoral Researcher with the University of New South Wales, Canberra, ACT, Australia, and then, he worked as a Postdoctoral Researcher with the Department of Physics, National Cheng Kung University, Tainan City, Taiwan. In July 2017, he joined Shanghai Jiao Tong University. His research interests include quantum control, optimization, and intelligent control of complex systems. 
\end{IEEEbiography}

\begin{IEEEbiography}[{\includegraphics[width=1in,height=1.25in,clip,keepaspectratio]{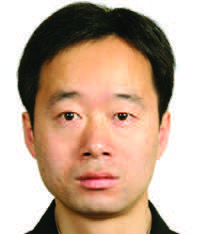}}]{Dewei Li}
received the B.S. degree and the Ph.D. degree in automation from Shanghai Jiao Tong University, Shanghai, P.R. China, in 1993 and 2009, respectively.

He is a Professor with the Department of Automation at Shanghai Jiao Tong University. He worked as a Postdoctoral Researcher with Shanghai Jiao Tong University from 2009 to 2010. His research interests include model predictive control, intelligent control, and intelligent transportation systems.
\end{IEEEbiography}

\end{document}